\documentclass{article}

\usepackage[final]{nips_2016}

\usepackage[utf8]{inputenc} 
\usepackage[T1]{fontenc}    
\usepackage{hyperref}       
\usepackage{url}            
\usepackage{booktabs}       
\usepackage{amsfonts}       
\usepackage{nicefrac}       
\usepackage{microtype}      
\usepackage{amsmath}
\usepackage[export]{adjustbox}
\usepackage{algpseudocode}
\usepackage{graphicx}
\usepackage{xargs}                      
\usepackage[pdftex,dvipsnames]{xcolor}  
\usepackage[colorinlistoftodos,prependcaption,textsize=tiny]{todonotes}
\newcommandx{\unsure}[2][1=]{\todo[linecolor=red,backgroundcolor=red!25,bordercolor=red,#1]{#2}}
\newcommandx{\change}[2][1=]{\todo[linecolor=blue,backgroundcolor=blue!25,bordercolor=blue,#1]{#2}}
\newcommandx{\info}[2][1=]{\todo[linecolor=OliveGreen,backgroundcolor=OliveGreen!25,bordercolor=OliveGreen,#1]{#2}}
\newcommandx{\improvement}[2][1=]{\todo[linecolor=Plum,backgroundcolor=Plum!25,bordercolor=Plum,#1]{#2}}
\newcommandx{\thiswillnotshow}[2][1=]{\todo[disable,#1]{#2}}
\usepackage{bm}
\usepackage{caption}
\usepackage{subcaption}
\usepackage{graphicx}

\graphicspath{ {figures/} }

\title{Bayesian Nonparametric Modeling of Heterogeneous Groups of Censored Data}
\author{
  Alexandre Pich\'{e} \\
  Department of Mathematics and Statistics \\
  McGill University \\
  \texttt{alexandre.piche@mail.mcgill.ca} \\
   \And
   Russell J. Steele\\
  Department of Mathematics and Statistics \\
  McGill University \\
  \texttt{russell.steele@mcgill.ca} \\
   \AND
   Ian Shrier \\
   Lady Davis Institute for Medical Research \\
   McGill University \\
   \texttt{ian.shrier@mcgill.ca} \\
   \And
   Stephanie Long \\
   Department of Family Medicine\\
   McGill University\\
   \texttt{stephanie.long@mail.mcgill.ca} \\
}

\begin{document}

\maketitle
\begin{abstract}

Datasets containing large samples of time-to-event data arising from several
small heterogeneous groups are commonly encountered in statistics. This presents
problems as they cannot be pooled directly due to their heterogeneity or
analyzed individually because of their small sample size. 
  Bayesian nonparametric modelling approaches can be
  used to model such datasets given their ability to flexibly share information
  across groups. In this paper, we will compare three popular Bayesian
  nonparametric methods for modelling the survival functions of heterogeneous
  groups. Specifically, we will first compare the modelling accuracy of the
  Dirichlet process, the hierarchical Dirichlet process, and the nested
  Dirichlet process on simulated datasets of different sizes, where group
  survival curves differ in shape or in expectation. We, then, will compare the
  models on a real-world injury dataset. 
\end{abstract}

\section{Introduction}

Survival analysis is widely used in healthcare to model time-to-injury occurrence. Often, the data are stratified in multiple small groups that differ
in certain aspects, such as exposure or risk, and cannot be directly pooled to
estimate the survival function. Given that the numbers of injuries are often
rather small relative to exposure time, and that most individuals do not
experience injury during follow-up, it is useful to introduce dependence across
group parameters to obtain precise estimates of the injury rates. Bayesian
nonparametric methods such as the Dirichlet process (DP)
\citep{ferguson1973bayesian} and Gaussian process
have been widely used in survival analysis, but, surprisingly popular extensions
of the DP have yet to be explored in this context.
Specifically, we adapted the nested Dirichlet process (NDP) 
\citep{rodriguez2012nested} and hierarchical Dirichlet process (HDP)
\citep{teh2012hierarchical} to model the survival functions of related groups,
in which some data has been censored. 

The shape of the survival function provides information that cannot be obtained
from the parameters alone. In practice, survival functions are unlikely to follow a
parametric family and could be better modelled by a mixture of parametric
distributions \citep{Farewell1982}.  Thus, the DP is well-suited to model the
survival function since it makes weak assumptions about the shape of the curve,
yet provides smooth estimates \citep{blum1977, kuo1997bayesian,
  kottas2006nonparametric}.  

In this paper, we are particularly interested in the case in which the dataset is
composed of multiple groups, where each group is a mixture with an unknown
number of components, and some components' parameters might be identical across
groups. To take advantage of the structure of such dataset, sharing information
at the observation and the group level is necessary. We will use the term
'groups' to refer the structure present in the data, and 'clusters' as the structure
inferred by the model. 

\section{Adaptation of the Models to Handle Censored Observations}

We are interested in modelling a collection of survival functions, where the
$j$-th group's survival function is denoted as $S_{j}(t) = 1 - F_j(t) = P(X_j >
t)$. The $i$-th observation of the $j$-th group is denoted as $Y_{ji} = \min
\{X_{ji}, Z_{ji}\}$, where $Z_{ji}$ and $X_{ji}$ represent the censoring
time and event time, respectively. These are mutually independent
random variables. In the case where $Z_{ji} < X_{ji}$, we say that the
observation has been censored and we let the indicator variable $\gamma_{ji}$ be
equal to $0$, and to 1 otherwise.

The case $Z_{ji} < X_{ji}$ indicates that the observation has been censored and
we let the indicator variable $\gamma_{ji}$ be equal to $0$, and to 1 otherwise.
 
Censoring makes estimation challenging, because it masks important information
about $P(X_j > t)$. Fortunately, the DP can be modified to handle censored
observations by: (1) including the survival function in the
likelihood; and (2) using a Gibbs sampling step to augment censored
observations. 

\subsection{Likelihood Function}

The likelihood function is used to assign observations to mixture components.
Specifically, we assigned the observation $ji$ to the component $l$ with
probability: $p(\xi_{ji}=l) \propto w_{l} f(y_{ji}|\theta_{l})$, where $w_{l}$
is the weight of the $l$-th atom obtain via stick breaking
\citep{sethuraman1994constructive}, and $f(y_{ji}|\theta_{l})$ is the density of
the distribution with parameter $\theta_l$ evaluated at $y_{ji}$. 

To handle censored observations, we will modify the probability of sampling an
indicator variable to: $p(\xi_{ji}=l) \propto w_{l}[ \gamma_{ji}
f(y_{ji}|\theta_{l}) + (1-\gamma_{ji}) S(y_{ji}|\theta_l)]$, where
$S(y_{ji}|\theta_l)$ is the survival function of the distribution with
parameter $\theta_l$ evaluated at $y_{ji}$, and $\gamma_{ji}$ is an indicator
variable equal to 0 if the observation $ji$ has been censored, and 1 otherwise.
This is intuitive since the survival function is the probability of surviving
past time $y_{ji}$. This modification of the likelihood can easily be
performed for the HDP. 

This can also be applied to the NDP by assigning group $j$ to the mixture of
distributions $k$ with probability $p(\zeta_j=k) \propto \pi_k \prod_{i=1}^{I_j}
\sum_{l=1}^L w_{lk} p(y_{ji}|\theta_{lk})$, where $I_j$ is the number of
observations in group $j$, and $\pi_k$ is a weight obtained via the stick
breaking construction. \citep{rodriguez2012nested} 

\subsection{Gibbs Step}

Once a censored observation has been assigned to cluster $l$, it cannot be
naively used to update the parameter's posterior, neither can we ignore the
censored observation, since it would bias our estimates. Instead, we can use a
Gibbs step to simulate what the observation could have been if it was not
censored at time $y_{ji}$ given that it belongs to cluster $l$. 

Specifically, censored observation $ji$ assigned to cluster $l$ can be filled by
drawing a value from $\hat{y}_{ji} \sim p(\theta_l|\hat{y}_{ji}>y_{ji})$. We can
use these simulated values to obtain an unbiased estimate of our parameters. It is
straightforward to modify the NDP and HDP algorithm in a similar fashion. 

\section{Results}

We focused on datasets containing small heterogeneous groups of mixtures, where
the number of mixtures are unknown and a high proportion of observations are
censored. In this case, the groups could not be pooled directly without making
strong assumptions about their homogeneity. Further, due to the small sample
size it was impractical to fit a different survival curve to each group. Fitting
a proportional hazard model would also be difficult for the same reasons.
Instead, we took advantage of the BNP methods’ ability to infer clusters from
the data, and therefore, estimated their survival function more efficiently. To
model the data using BNP, we assumed that the observations were exchangeable
within groups and across groups. For instance, observations within groups could
be relabeled and groups themselves could also be relabeled.  

We will compare the BNP methods to a Weibull Gamma frailty model (GFM) fitted
using the parfm R package \citep{munda2012parfm}. The GFM confidence interval has
been computed using the inverse of the negative hessian as an approximation
of the parameters' covariance matrix.

\subsection{Simulation Study}

This simulation study aims to demonstrate the statistical pooling power of
different models in a commonly occurring situation.  The simulated task requires
modelling survival functions in an injury dataset that is characterized by high
levels of censoring and small, but, numerous heterogeneous groups. To replicate
these characteristics, we simulated 600 observations from three
mixtures with a censoring rate of approximately 50\% using the survsim
package \citep{survsim}. The parameters and weights of the 
mixtures were chosen so that the curves from mixture one and mixture two had
similar shapes, but different expectations, and the curves from mixture one and
mixture three had similar expectations, but different shapes, as seen in
\textbf{Figure 1}. Furthermore, each group has its own frailty parameter
originating from a $Gamma(1,1)$. 

We will compare the methods on different structures of the dataset e.g. varying
the number of groups $J$ and the group size $n_j$ while keeping the overall
sample size fixed at 600. To quantify the accuracy of our model, we computed the
mean log pointwise predictive density of our model. In addition, to evaluate
uncertainty we computed the mean width of our 95\% credible interval and the
proportion 
captured by the real survival curve. The results are reported in \textbf{Figure
  2} and \textbf{Figure 3}.

\begin{center}
  \includegraphics[scale=0.4]{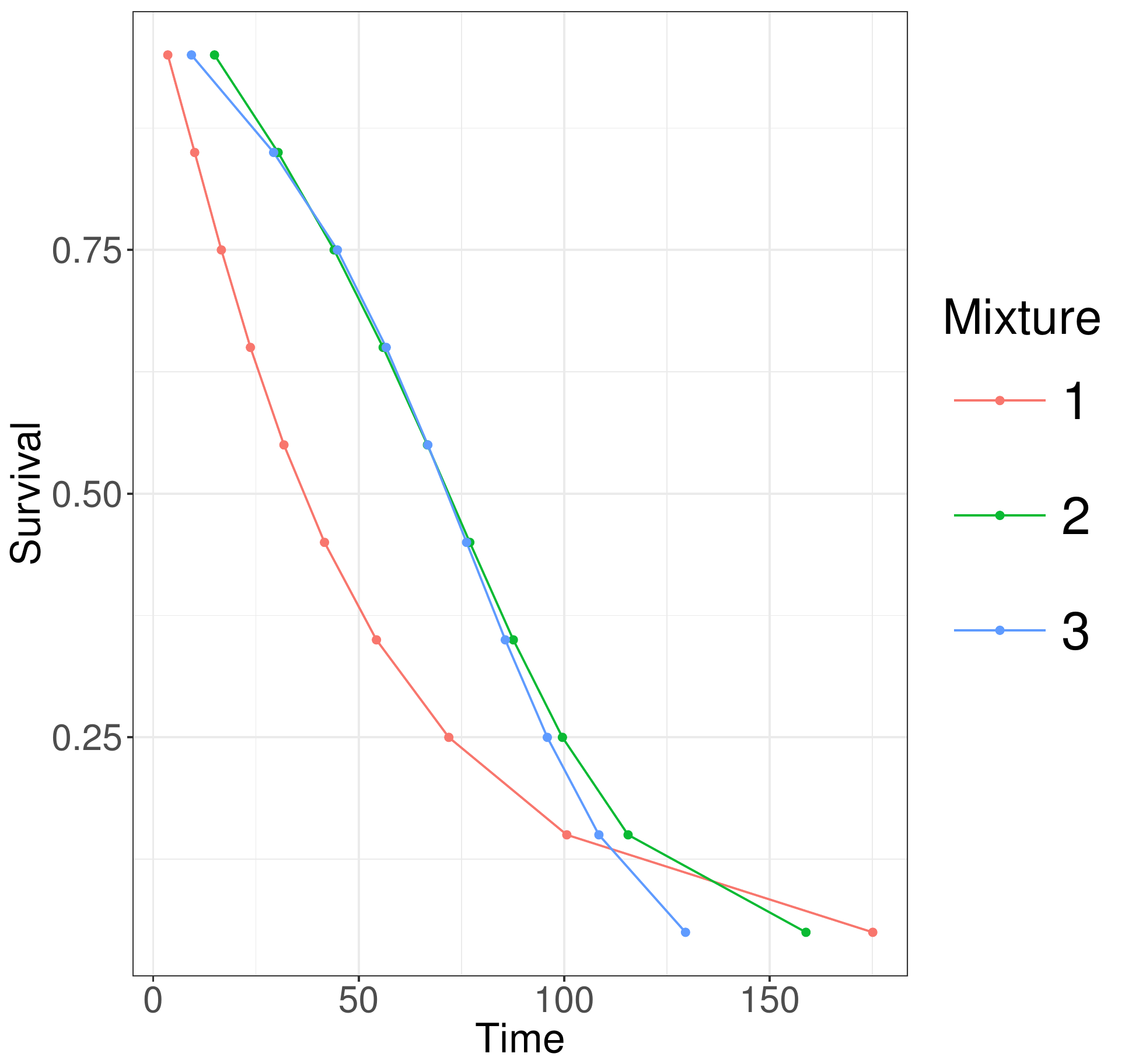}
  \captionof{figure}{This displays the mixtures' parameters and weights that were chosen for this simulation. The curves for mixture 1 and mixture 2 have similar shapes, but differ in expectations, and the curves for mixture 1 and 3 have similar expectation, but different shapes.}
\label{fig:simDataset}
\end{center}

\subsection{Performance Arts Company}

Based on the successful modelling of the synthetic data, the model was applied to
a performance art company dataset to determine its performance in real world
applications. Similar to the simulation study, this dataset consists of several,
clearly related, but heterogeneous groups that cannot be directly pooled without
making strong assumptions.  

The dataset was structured into shows and roles, both of which could be used as
groups. However, it was more reasonable to group the observations by shows, than
by
roles, because roles vary within shows but also vary across shows in terms of
exposure and risk. This resulted in 279 unique groups with a group size ranging
from 1 to 69 observations. An overview of the dataset can be seen in
the Appendix.  

\subsection{Discussion}

\begin{center}
\includegraphics[width=0.4\linewidth]{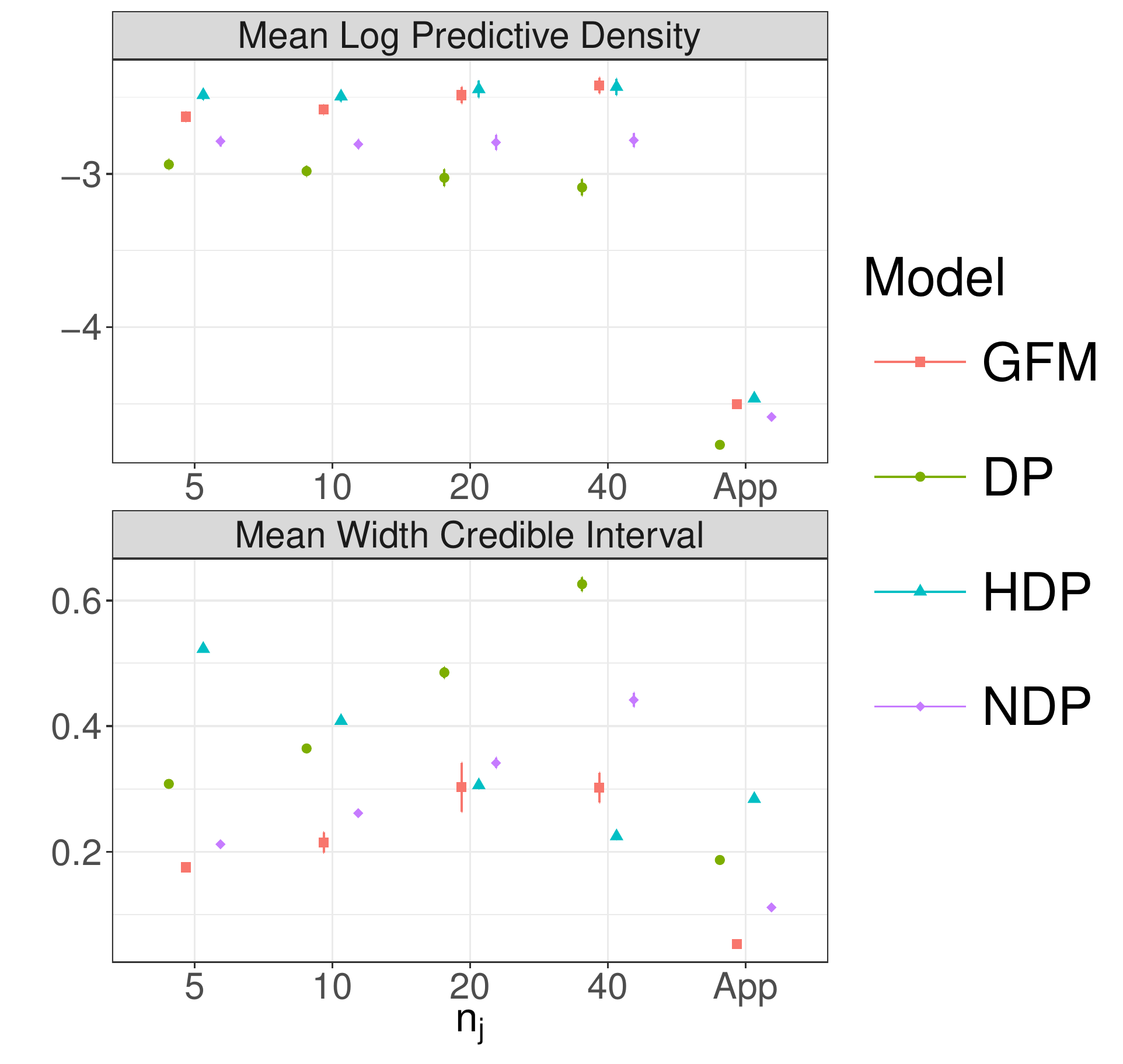}
\captionof{figure}{The mean log pointwise predictive density is reported for
  different structures of the dataset. Note that a higher mean log predictive
  density implies better modelling of the dataset. The mean width credible
  interval is also reported in the second panel.}
\end{center}

\begin{center}
\includegraphics[width=0.4\linewidth]{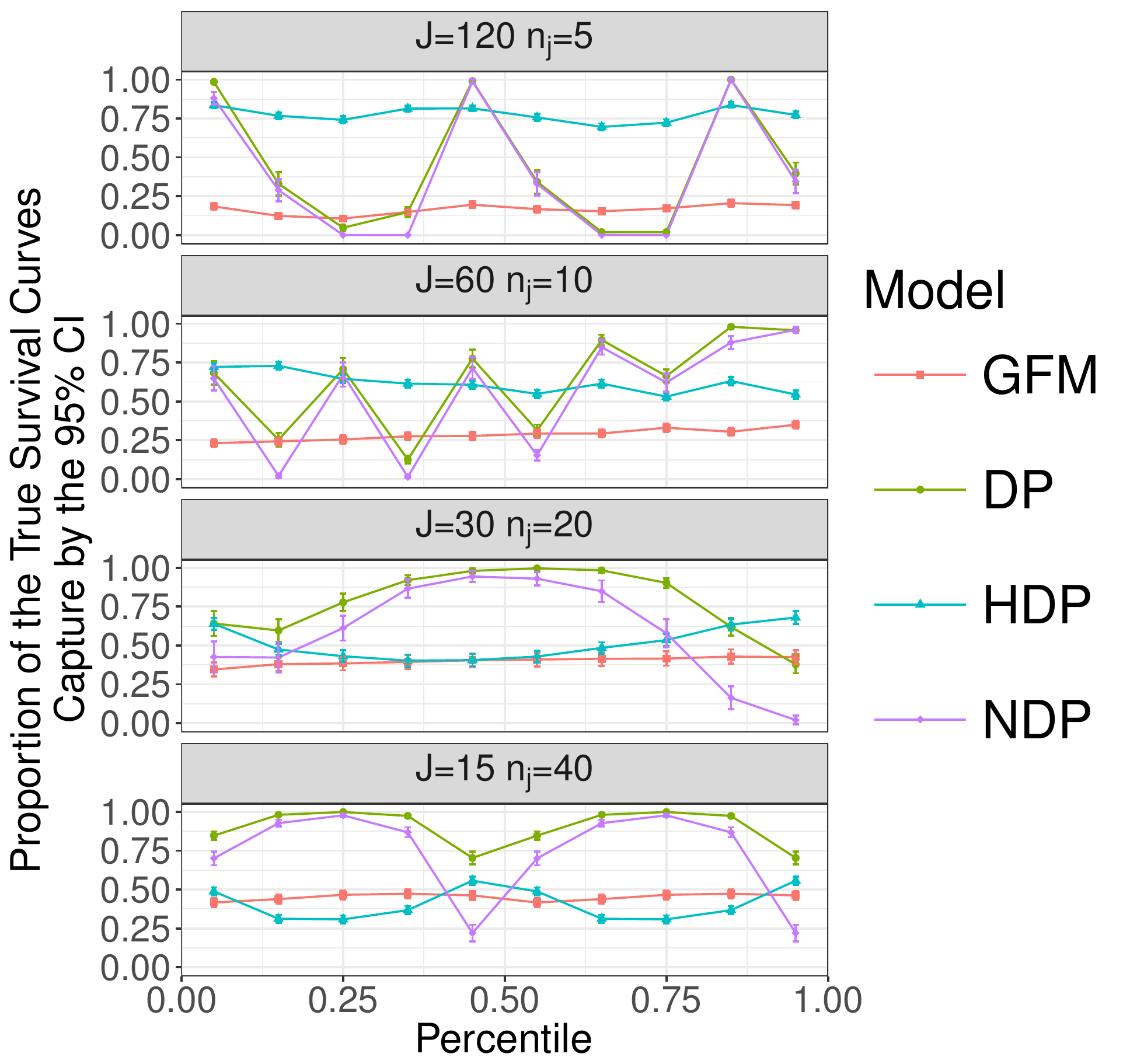}
\captionof{figure}{The average proportion of coverage of the true survival
  curves have been computed at different percentiles of the 3 mixture of survival
curves. The percentiles are the same as the identified in \textbf{Figure 1}.}
\end{center}

From the results, the BNP methods' performance depends on
the structure of the dataset. On the one hand, based on the mean log predictive
density it seems like the HDP is the most adequate method to model small
heterogeneous groups, and the complex performance art dataset's survival curves.
On the other hand, the NDP shows the ability to recover the original survival 
curves while keeping a reasonable credible interval width.

\section{Conclusion}

This paper compared three different BNP methods on a predictive task in a
survival analysis context. First, they were compared on a simulated dataset
with groups that came from three mixtures. Following, they were compared on a
real world injury dataset from a 
Performance Art Company. For each of these analyses, the BNP models were a
reasonable choice given that the groups could not be pooled directly, but were
intuitively related. Using BNP methods, the clusters could be directly inferred
from the data, and certain groups and observations could be pooled together to
obtain better estimates of the survival curves. 

It would be beneficial for the analysis of injury data, if BNP methods were used
more often since they provide a good way to deal with uncertainty and allow for
sharing of information across groups with high censoring. The next step would be
to extend 
our model to handle recurrent events, as they are highly common in the analysis
of injury data.

\subsubsection*{Acknowledgments}
The work has been supported by CIHR and NSERC. The authors would like to thank James
McVittie, Meng Zhao and all the members of professors Shrier and Steele's group
for their helpful comments.
\small

\bibliography{../../Bibliography}

\section*{Appendix}

\begin{figure}[h]
  \includegraphics[scale=0.55]{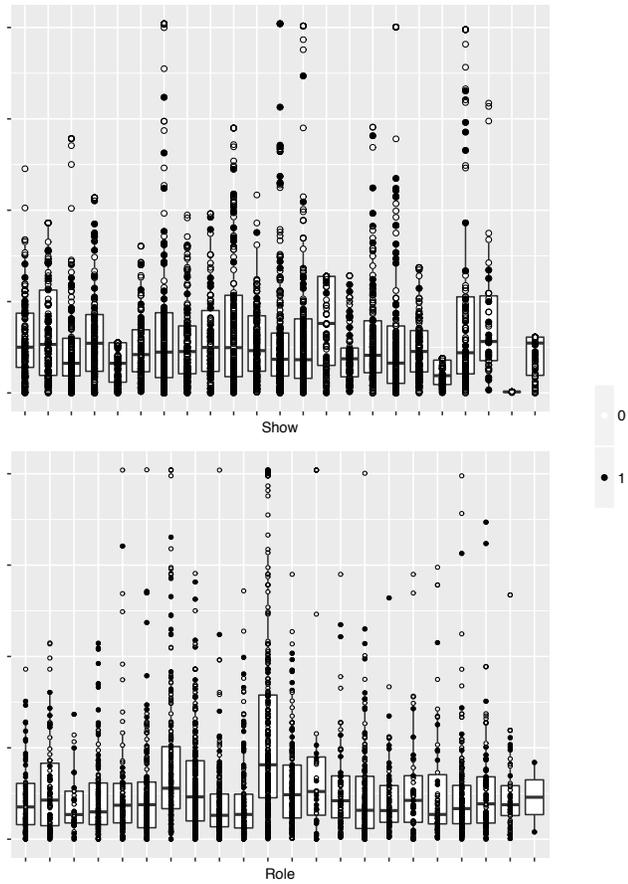}
\caption{Overview of the Performance Art Company dataset which was first grouped by show and then by role, whereby the white dots indicate censored observations and the black dots indicate time to first injury.}
\label{fig:paDataset}
\end{figure}

\end{document}